\pdfoutput=1

\documentclass[11pt]{article}

\usepackage[final]{coling}

\usepackage{times}
\usepackage{latexsym}

\usepackage[T1]{fontenc}

\usepackage[utf8]{inputenc}

\usepackage{microtype}

\usepackage{inconsolata}

\usepackage{graphicx}

\usepackage{booktabs}
\usepackage[most]{tcolorbox}
\tcbuselibrary{fitting,skins}
\usepackage{listings}
\newtcblisting{tcbverbatim}{
  fontupper=\tiny,
  blank,
  borderline={1pt}{-2pt},
  nobeforeafter,
  boxsep=1pt,top=0pt,bottom=0pt,
  listing only,
  listing options={
    language=python,
    basicstyle=\scriptsize\ttfamily,
    columns=flexible,
    breaklines,
  }
}
\usepackage{fancyvrb}
\usepackage{hyperref} 
\usepackage{multirow} 
\usepackage{svg}
\usepackage{subcaption}
\title{Multimodal Fact-Checking with Vision Language Models: A 
Probing Classifier based Solution with Embedding Strategies
}

\author{
  \textbf{Recep Firat Cekinel\textsuperscript{1}},
  \textbf{Pinar Karagoz\textsuperscript{1}},
  \textbf{Çağrı Çöltekin\textsuperscript{2}},
\\
\\
 \textsuperscript{1}Middle East Technical University, Turkiye
  \textsuperscript{2}University of Tübingen, Germany
\\
  \small{
    \textbf{Correspondence:} \href{mailto:rfcekinel@ceng.metu.edu.tr}{rfcekinel@ceng.metu.edu.tr}
    }
}

\begin{document}
\maketitle
\begin{abstract}

This study evaluates the effectiveness of Vision Language Models (VLMs) in representing and utilizing multimodal content for fact-checking. To be more specific, we investigate whether incorporating multimodal content improves performance compared to text-only models and how well VLMs utilize text and image information to enhance misinformation detection. Furthermore we propose a probing classifier based solution using VLMs. Our approach extracts embeddings from the last hidden layer of selected VLMs and inputs them into a neural probing classifier for multi-class veracity classification. Through a series of experiments on two fact-checking datasets, we demonstrate that while multimodality can enhance performance, fusing separate embeddings from text and image encoders yielded superior results compared to using VLM embeddings. Furthermore, the proposed neural classifier significantly outperformed KNN and SVM baselines in leveraging extracted embeddings, highlighting its effectiveness for multimodal fact-checking.

\end{abstract}

\section{Introduction}
\label{sec:intro}

\begin{figure}[t!]
    \centering
\includegraphics[scale=0.4]{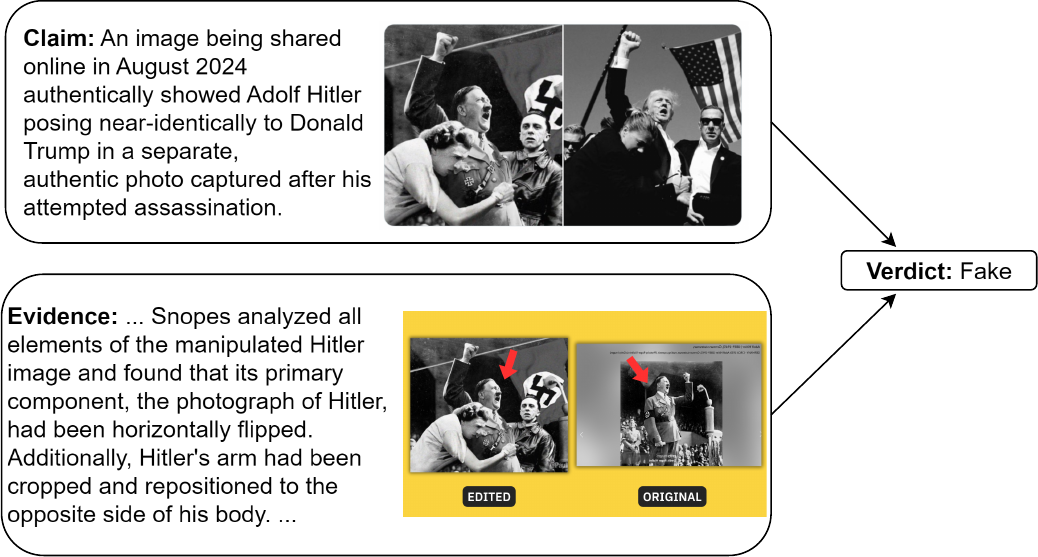}
    \caption{Example multimodal fact-checking from Snopes}
    \label{fig:mfc_intro}
\end{figure}

Social media platforms are increasingly becoming the primary source of news for many people. However, these platforms are susceptible to the rapid spread of fake stories, which can be used to manipulate public opinion \cite{allcott2017social}. Fabricated posts may include false text, images, videos, or speech content \cite{alam2022survey, akhtar2023multimodal, comito2023multimodal}, designed to deceive social media users. Therefore, automated fact-checking systems should be able to consider information from different modalities \cite{abdali2024multimodal}. For instance, on the Snopes website, a claim\footnote{\url{https://www.snopes.com/fact-check/hitler-trump-image-fake/}} about an edited image was proven to be fake by providing the original image and explaining how it was fabricated to manipulate public opinion about public figures. To verify the truthfulness of such content, it is essential to process both text and image information (see Figure \ref{fig:mfc_intro}).

A vision language model (VLM) consists of an image encoder, a text encoder and a mechanism such as contrastive learning \cite{bordes2024introduction} and cross attention \cite{chen2022visualgpt} to fuse text and image information. By this way, the model leverages the text and visual information while generating a response text. VLMs consist of billions of parameters and fine-tuning these models requires significant computational resources. Although parameter-efficient fine-tuning approaches \cite{hu2022lora, liu2024dora} have proven to be very effective for large language models, VLMs do not scale well horizontally. Consequently, such VLMs cannot be fine-tuned with moderate batch size and sequence length on a single GPU for problems like fact-checking that requires long text inputs.

Instead of fine-tuning, probing classifiers are trained on the representations of a pre-trained model \cite{kunz-kuhlmann-2020-classifier} to predict linguistic features such as dependency parsing \cite{adelmann2021impact} and POS tagging \cite{kunz2021test}. A key advantage of probing classifiers is their ability to assess how well the pre-trained model has captured linguistic properties. In this study, we aim to evaluate how VLMs leverage both text and images for the fact-checking task by training a probing classifier. The following research questions are addressed in the paper.

\textbf{RQ1: Validating the need for multimodality:} Does incorporating multimodal data improve performance in the fact-checking task or are text-only models sufficient?

\textbf{RQ2: Leveraging multimodal content:} How effectively do VLMs utilize both text and image information to enhance fact-checking performance?

\textbf{RQ3: Evaluating probing classifiers:} How does a probing neural classifier compare to baseline models in the context of the fact-checking task?

This study proposes a probing classifier that involves extracting the last hidden layer's representation and using it as input for a neural network. By introducing this pipeline, we aim to elaborate on the utilization of multimodal information, text and image, compared to embeddings extracted from discrete text-only and image-only models for the fact-checking problem. The source code is available at the following anonymous \href{https://github.com/firatcekinel/Multimodal-Fact-Checking-with-Vision-Language-Models}{GitHub repository}\footnote{https://github.com/firatcekinel/Multimodal-Fact-Checking-with-Vision-Language-Models}

\section{Related Work}

\paragraph{Text-Based Fact-Checking}

Shared tasks such as FEVER \cite{thorne2018fever}, CLEF2018 \cite{clef2018checkthat:overall} and AVeriTeC \cite{NEURIPS2023_cd86a305} evaluate fact-checking systems on textual claims. Although LLMs achieved high success rates on fact-checking with English data even in zero-shot settings \cite{hoes2023leveraging}, \citet{zhang2024we} emphasize the need for language models that are specifically pre-trained on the target language. Similarly \citet{cekinel-etal-2024-cross-lingual} investigate cross-lingual transfer learning using LLMs. Additionally, \citet{cheung2023factllama} incorporate external evidence during instruction-tuning to enhance the knowledge of LLMs. Moreover, \citet{yue2023metaadapt} focus on cross-domain knowledge transfer with in-context learning. \citet{tang2024minicheck}  verify the factuality of synthetically generated claims against grounding documents. LLMs are also used for explanation generation \cite{bangerter2024hybrid, zeng2024justilm, mediratta2024enabling} and neuro-symbolic program generation \cite{pan-etal-2023-fact} for fact-checking. While these works primarily focus on enhancing models' knowledge, we aim to explore how they can leverage different modalities. 

\paragraph{Multimodal Fact-Checking}

While SpotFake+ \cite{singhal2020spotfake+} concatenates extracted text and image features for further processing through feed-forward layers, CARMN \cite{song2021multimodal} fuses multimodal information using a cross-modal attention residual network. Pre-CoFactv2 \cite{du2023team} implements a multi-type fusion model that uses cross-modality and cross-type relations. COOLANT \cite{wang2023cross} implemented a contrastive learning based fusion method for image-text alignment. \citet{gao2024knowledge} incorporates the information extracted from the tweet graph with text and image embeddings for improving fake news detection. \citet{liu2024exploring} examined the impact of audio in multimodal fact-checking by proposing a framework that fuses text, video and audio information with the cross-attention mechanism. \citet{wang2024fake} align news text with images by cross-modal attention model.

\citet{geng2024multimodal} propose an evaluation framework for VLMs that assesses the pre-trained knowledge of these models in fact-checking without evidence. RAGAR \cite{khaliq2024ragar} presents a RAG-based model that reframes the problem as question-answering for retrieved evidence pieces. MMIDR \cite{wang2024mmidr} trains a distilled model to generate explanations. SARD framework \cite{yan2024sard} applies multimodal semantic alignment to integrate multimodal network features.  LVLM4FV \cite{tahmasebi2024multimodal} is an evidence-ranking approach and was evaluated on two benchmark datasets using LLMs and VLMs with zero-shot setting.


Although recent studies have focused on developing multimodal models for fact-checking using various fusion approaches, we aim to explore how effectively VLMs utilize different modalities. \citet{geng2024multimodal} also evaluated the robustness of recent VLMs for this problem by comparing the pre-trained knowledge of selected models and their prediction accuracy and confidence rates in zero-shot and few-shot settings. In contrast, we aim to leverage VLM representations by proposing a pipeline that trains a classifier using these embeddings. Furthermore, our primary focus is on utilizing multimodal information. In the experiments, we evaluate the intrinsic fusion of multimodal information against the extrinsic fusion of separate text-only and image-only representations.
\begin{figure*}[th!]
    \centering
\includegraphics[width=15cm, height=9.5cm]
{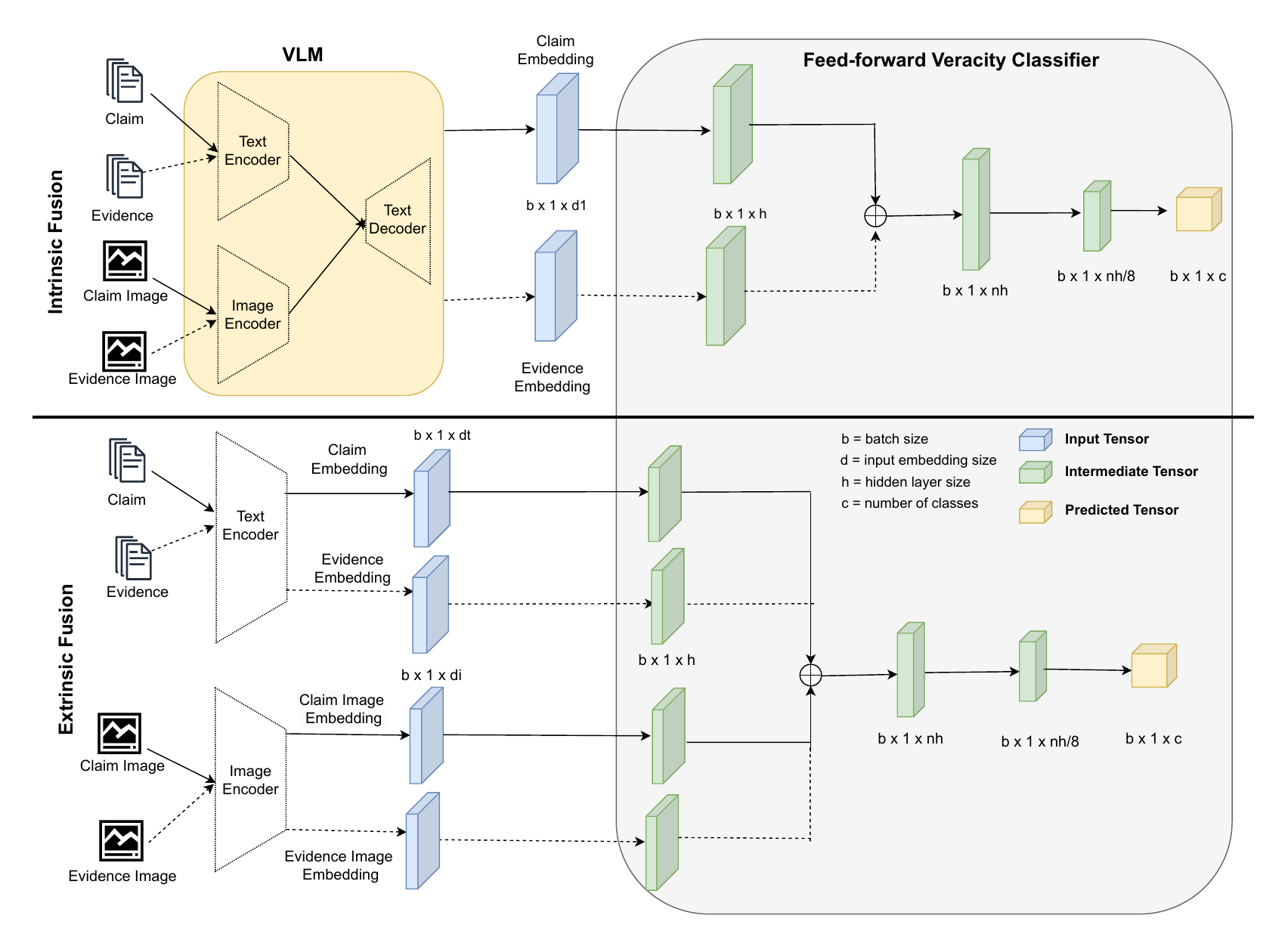}
    \caption{Overview of our probing fact-verification classifier. ReLU activation is applied after each linear layer with dropout for better generalization. The dashed lines indicate optional embeddings. In other words, evidence text and evidence image representations are optional in this pipeline.}
    \label{fig:ffclassifier}
\end{figure*}

\section{The Proposed Method}

\subsection{Feed-Forward Veracity Classifier}
\label{sec:probing_classifier}

We introduce a probing classifier to examine the efficiency of multimodal embeddings compared to separate embeddings extracted from text-only and image-only models for veracity prediction. The VLM embeddings fuse text and image modalities intrinsically but distinct text and image encoder embeddings are fused extrinsically by the probing classifier as illustrated in Figure \ref{fig:ffclassifier}.

First, the last hidden layer representation is extracted from a VLM or a text/image encoder. The neural classifier either receives the VLM representation or embeddings from the corresponding text encoder and image encoder, then predicts veracity classes. If multiple input tensors are fed to the neural classifier, they are processed by a linear layer and after the first layer, all tensors are resized to a "hidden\_size" — a hyper-parameter determined by validation experiments — and then concatenated. We concatenate after the first layer because the text and image embedding sizes vary significantly. To utilize both types of information equally, we resize these embeddings to the same dimension and concatenate them afterward. On the other hand, if only the VLM embedding is given to the network as input, two linear layers process the tensor sequentially without any concatenation.

In both of the probing classifier architectures, we implement a weighted cross-entropy loss, with weights determined by inverse class ratios to penalize the majority class more. Since PyTorch's cross-entropy loss implementation combines softmax with negative log-likelihood loss, the output tensor predicts class probabilities. Consequently, the classifier predicts the class with the highest probability for a given instance.

\subsection{Models}
\label{sec:models}

The primary goal of this study is to examine whether merging image and text information provides gains for the fact-checking problem. To this end, we selected three multimodal models with different fusion mechanisms, as explained below.

\textbf{Qwen-VL} \cite{Qwen-VL} is a multimodal model introduced by Alibaba Cloud. Qwen-VL is based on the Qwen-7B \cite{bai2023qwen} language model and Openclip's ViT-bigG \cite{ilharco_gabriel_2021_5143773} vision transformer. The model leverages both modalities through a cross-attention mechanism. Information from the vision encoder is fused into the language model using a single-layer cross-attention adapter with query embeddings optimized during the training phase. In this study, we employed \textit{Qwen-VL-Chat-Int4} checkpoint which was the 4-bit quantized version.

\textbf{Idefics2} \cite{laurençon2024matters} is a general-purpose multimodal VLM introduced by Huggingface. It is based on the Mistral-7B \cite{jiang2023mistral} language model and SigLIP's vision encoder \cite{zhai2023sigmoid} (SigLIP-So400m/14). The model employs a vision-language connector that takes the vision encoder's representation as input, using perceiver pooling and MLP modality projection. After these operations, the image information is concatenated with the encoded text representation and fed into the language model decoder.

\textbf{PaliGemma} \cite{beyer2024paligemma} is introduced by Google and is based on the Gemma-2B \cite{team2024gemma} language model and SigLIP's vision encoder \cite{zhai2023sigmoid} (SigLIP-So400m/14). Since Gemma-2B is a decoder-only language model, the vision encoder's representation is fed into a linear projection, concatenated with text inputs, and then fed into the Gemma-2B language model for text generation. In this study, we employed \textit{paligemma-3b-mix-448} checkpoint that was fine-tuned on a mixture of downstream tasks.

\subsection{Datasets}

\textbf{Mocheg} \cite{yao2023end} consists of 15K fact-checked claims from Politifact\footnote{\url{https://www.politifact.com/}} and Snopes.\footnote{\url{https://www.snopes.com/}} These websites employ journalists to verify claims who collect evidence documents and write ruling comments. The Mocheg dataset includes both text and image evidence which were crawled from the reference articles linked on the fact-checked claims' webpages. In cases where multiple evidence images were available for a claim, some collected images were found to be irrelevant. Therefore, for the experiments, only the first image was used as the evidence image.

\textbf{Factify2} \cite{suryavardan2022factify} is a challenge dataset containing 50K claims. The authors collected true claims from tweets by Indian and US news agencies and false claims from fact-checking websites. They scraped text and image evidence from external articles and also collected claim images from the headlines of the claims. The fact-verification task was reformulated as an entailment problem where claims were annotated to indicate whether the claim text and image were entailed by the evidence text and image.

\section{Experiments}
\label{sec:experiments}
We conducted experiments on compute nodes with 4x40GB Nvidia A100 GPUs. While evaluating the models on the datasets, we ignore the instances that have missing text evidence or images. For the Mocheg dataset, we used the original train-dev-test splits. The dataset has three labels \textit{"supported"}, \textit{"refuted"} and \textit{"not enough info (NEI)"} and we used the labels as it is.

Regarding the Factify2 dataset, since the labels in the test set were unavailable, the original validation data was kept for testing. Instead, we randomly selected 10\% of the training set for validation but kept the same percentages of classes in each split. Similar to \cite{tahmasebi2024multimodal}, we reduced the original five labels to three classes:  \textit{Support} (Support\textunderscore Multimodal \& Support\textunderscore Text), \textit{Refute} and \textit{Not enough info} (Insufficient\textunderscore Multimodal \& Insufficient\textunderscore Text) to evaluate the proposed approach.

During the training of the probing classifier using the embeddings, validation experiments were conducted through grid search within the parameter space detailed below. Note that only the best parameter settings are presented in Appendix \ref{sec:appendix_hyperparam}. Last but not least, we reported F1-macro scores and F1 scores for each class in the following experiments.

\subsection{Zero-Shot Inference}
\label{sec:exp_inference}
\begin{table*}[th!]
\centering
\resizebox{\linewidth}{!}{
\begin{tabular}{ll|rrrr|rrrr}
\multicolumn{1}{c}{}               & \multicolumn{1}{c}{}                & \multicolumn{4}{c}{MOCHEG}                                                                                                                            & \multicolumn{4}{c}{FACTIFY2}                                                                                                                          \\
\hline
\multicolumn{1}{c}{\textbf{Models}} & \multicolumn{1}{c}{\textbf{Inputs}} & \multicolumn{1}{c}{\textbf{Support}} & \multicolumn{1}{c}{\textbf{Refute}} & \multicolumn{1}{c}{\textbf{NEI}} & \multicolumn{1}{c}{\textbf{F1-macro}} & \multicolumn{1}{c}{\textbf{Support}} & \multicolumn{1}{c}{\textbf{Refute}} & \multicolumn{1}{c}{\textbf{NEI}} & \multicolumn{1}{c}{\textbf{F1-macro}} \\
\hline
Qwen-7B        & text            & 0.533            & 0.262           & 0.169        & 0.321             & 0.524            & 0.458           & 0.281        & 0.421             \\
Mistral-7B     & text            & 0.505            & 0.281           & 0.216        & 0.334             & 0.575            & 0.561           & 0.093        & 0.409             \\
Gemma-2b       & text            & 0.610            & 0.462           & 0.315        & 0.462             & 0.562            & 0.119           & 0.083        & 0.255             \\
\hline
Qwen-VL        & text + image    & 0.168            & 0.472           & 0.186        & 0.275             & 0.463            & 0.460           & 0.369        & 0.431             \\
Idefics2-8b    & text + image    & \textbf{0.619}            & 0.547           & 0.385        & 0.517             & 0.586            & \textbf{0.644}           & 0.303        & 0.511             \\
PaliGemma-3b   & text + image    & 0.222            & 0.347           & 0.449        & 0.339             & 0.149            & 0.139           & 0.186        & 0.158   \\        
\hline
LVLM4FV & text                        &  0.575                                & 0.542                               &  0.439                            & 0.519                                 &  0.593                                & 0.581                               &  \textbf{0.560}                            & 0.578                                 \\
LVLM4FV & text + image                        &   0.578                                &  0.569                               &   \textbf{0.457}                            &  \textbf{0.535}                                 & \textbf{0.678}                       & 0.605                               & 0.508                            & \textbf{0.597}                                 \\
MOCHEG & text + image                        & 0.490   & \textbf{0.604}                               &  0.282     &  0.459                                &  0.547      & 0.621                              & 0.275                            & 0.481                                 \\
\end{tabular}
}
\caption{Text-only and multimodal inference results}
  \label{tab:vlm-inference}
\end{table*}

\begin{table*}[th!]
\centering
\resizebox{\linewidth}{!}{
\begin{tabular}{ll|rrrr|rrrr}
\multicolumn{1}{c}{}               & \multicolumn{1}{c}{}                & \multicolumn{4}{c}{MOCHEG}                                                                                                                            & \multicolumn{4}{c}{FACTIFY2}                                                                                                                          \\
\hline
\multicolumn{1}{c}{\textbf{Models}} & \multicolumn{1}{c}{\textbf{Inputs}} & \multicolumn{1}{c}{\textbf{Support}} & \multicolumn{1}{c}{\textbf{Refute}} & \multicolumn{1}{c}{\textbf{NEI}} & \multicolumn{1}{c}{\textbf{F1-macro}} &
\multicolumn{1}{c}{\textbf{Support}} & \multicolumn{1}{c}{\textbf{Refute}} & \multicolumn{1}{c}{\textbf{NEI}} & \multicolumn{1}{c}{\textbf{F1-macro}}\\
\hline
\multicolumn{1}{l}{PaliGemma-3b} & \multicolumn{1}{l}{text + image} & \multicolumn{1}{r}{0.412} & \multicolumn{1}{r}{0.514} & \multicolumn{1}{r}{0.173} & \multicolumn{1}{r}{0.366} &
 \multicolumn{1}{r}{ 0.751} & \multicolumn{1}{r}{0.997} & \multicolumn{1}{r}{0.757} & \multicolumn{1}{r}{0.835} \\
\end{tabular}
}
\caption{PaliGemma-3b fine-tuning results}
\label{tab:gemma_ft}
\end{table*}

In this experiment, we evaluated the zero-shot inference performance of text-only language models and multimodal VLMs on selected datasets. The text-only models were the same language models used in the VLMs for text processing. The purpose of reporting the results on text-only models is to examine the necessity of image content for the fact-checking problem.

For the text-only models, the claim and evidence text were provided as a single prompt, as illustrated in Figure \ref{fig:prompt}. Similarly, for each claim statement, the evidence text and evidence image were fed to the VLMs using a similar prompt template. Note that we reported results only for instances where the models responded with "supported," "refuted," or "not enough info." In other words, if the models did not provide a relevant justification, these cases were excluded from the reported results. 

\begin{figure}[th!]
\begin{Verbatim}[fontsize=\small]
Assess the factuality of the following claim by 
considering evidence. Only answer "supported", 
"refuted" or "not enough info".
Claim: {claim} 
Evidence: {evidence}
\end{Verbatim}
\caption{Prompt template}
\label{fig:prompt}
\end{figure}

We also reported the performance of two baseline models, LVLM4V \cite{tahmasebi2024multimodal} and MOCHEG \cite{yao2023end}, for comparison. MOCHEG concatenates the claim, evidence and image to generate CLIP \cite{radford2021learning} representations, employing attention mechanisms to update the claim representation based on the evidence. LVLM4V uses two-level prompting, formulating the problem as two binary questions and utilizing the Mistral \cite{jiang2023mistral} and LLaVa \cite{liu2024visual} models.

F1-macro scores along with F1 scores for each class are presented in Table \ref{tab:vlm-inference} for both text-only and multimodal models. The results show that multimodality can enhance performance depending on the dataset and model configuration. For example, both Idefics-8b and LVLM4FV consistently outperformed their text-only counterparts, while Qwen-VL performed slightly better on the Factify2 dataset but worse on the Mocheg dataset. In contrast, PaliGemma consistently responded with, "sorry, as a base VLM I am not trained to answer this question" to test queries, suggesting that specific policies were implemented in the base VLM to prevent responses to ambiguous queries. As a result, PaliGemma's inference performance was significantly lower than that of its language model counterpart, Gemma-2b (see Appendix \ref{sec:appendix_zs} for response frequencies). The inference scores of Idefics2-8b suggest that images may provide additional information for fact-checking, likely due to its fine-tuning on a mixture of supervised and instruction datasets, which could explain its success on these datasets. Additionally, LVLM4V’s prompting strategy appears more efficient, as it first checks whether the evidence is sufficient for verification before issuing a second prompt to verify or refute the claim.

\begin{figure*}[ht!]
    \centering
    \begin{subfigure}[b]{0.35\textwidth}
        \centering
        \includegraphics[width=\textwidth]{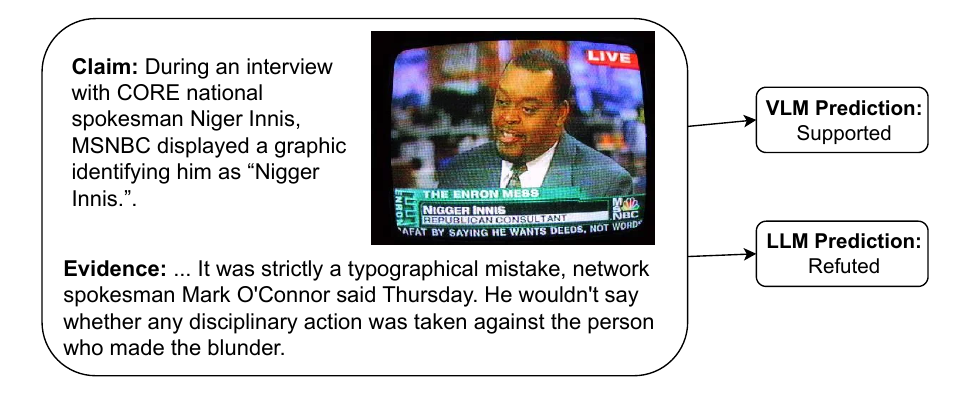}
        \caption{Supported claim}
        \label{fig:image1}
    \end{subfigure}
    \hfill
    \begin{subfigure}[b]{0.31\textwidth}
        \centering
        \includegraphics[width=\textwidth]{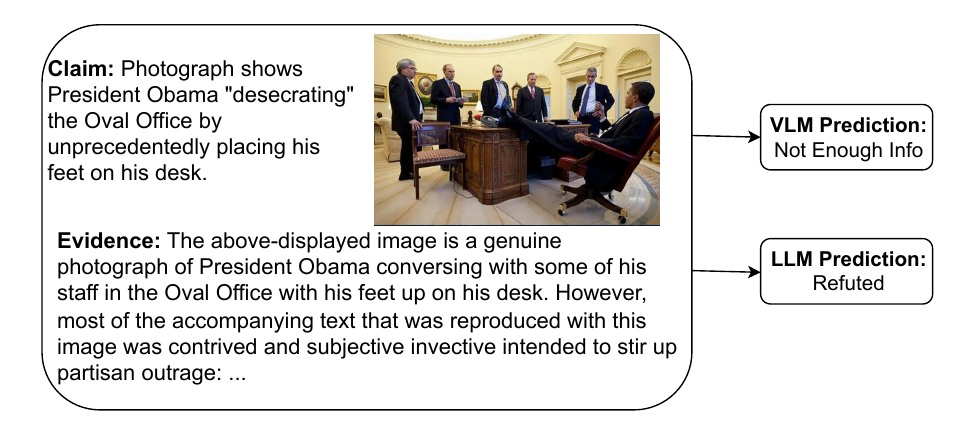}
        \caption{Unproven claim}
        \label{fig:image2}
    \end{subfigure}
    \hfill
    \begin{subfigure}[b]{0.31\textwidth}
        \centering
        \includegraphics[width=\textwidth]{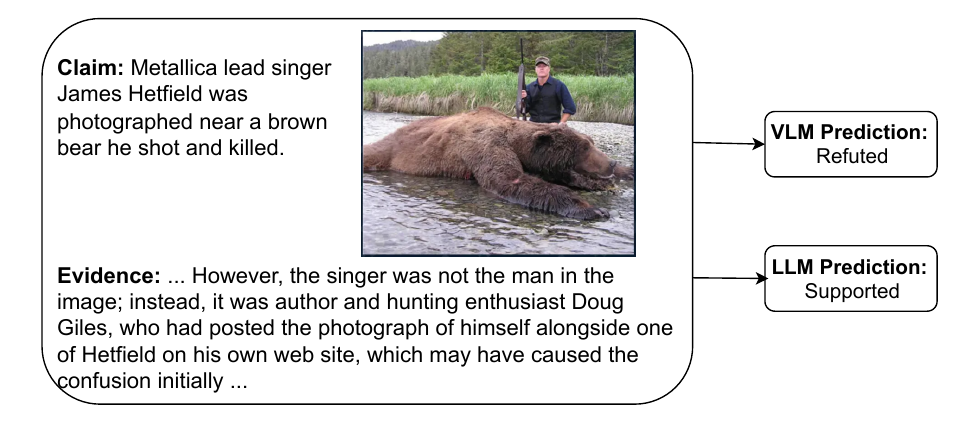}
        \caption{Refuted claim}
        \label{fig:image3}
    \end{subfigure}
    \caption{Qualitative examples for VLM and LLM inference predictions}
    \label{fig:qual_analysis}
\end{figure*}

\paragraph{Qualitative Analysis.} A qualitative analysis was conducted to explore the types of claims that were correctly predicted by multimodal models but incorrectly predicted by text-only models. In this analysis, the predictions from both the text-only (Mistral-7B) and multimodal (Idefics2-8b) models were employed on the Mocheg dataset. Although for the fact-checking problem, textual contents are the primary source, images are shown to be useful. After examining the instances that are correctly predicted by the VLM but misclassified by the LLM, we found that such instances required image information to accurately verify the claims, as illustrated in Figure \ref{fig:qual_analysis}. 

\paragraph{Fine-tuning PaliGemma-3b.} Fact-checking requires long evidence with supporting images, making it computationally challenging to fine-tune the VLMs with moderate batch sizes and sequence lengths on a single GPU. Therefore, we fine-tuned only the \textit{PaliGemma-3b-pt-224} checkpoint using claim, evidence and claim image as input. The experimental details are given in Appendix \ref{sec:appendix_ft}.

Evidence in the Mocheg dataset was collected from reference web articles. In contrast, Factify2 used the justifications provided by fact-checkers as evidence. As a result, Factify2's evidence is more concise and self-explanatory. However, models should interpret the knowledge from Mocheg's evidence sources to make a final decision. Because of the GPU memory considerations, evidence texts were cropped if they exceeded 768 words. 

Fine-tuning results, presented in Table \ref{tab:gemma_ft}, show a significantly lower score of 0.366 on the Mocheg dataset compared to inference results, due to cropping of the evidence text. However, on the Factify2 dataset, the evidence texts were shorter and the model leveraged the key information for making a decision and achieved 0.835 F1-macro score. Note that, on the Factify2 challenge the best-performing model was Logically \cite{gao2021logically} which was also fine-tuned on Factify2 dataset and it achieved 0.897 F1-macro score. Due to computational constraints, we were unable to utilize the long text evidence, particularly in the Mocheg dataset. As a result, we introduced a probing classifier instead of fine-tuning the selected VLMs.

\begin{table*}[th!]
\centering
\resizebox{\linewidth}{!}{
\begin{tabular}{ll|rrrr|rrrr}
                  &                 & \multicolumn{4}{c}{MOCHEG}                                              & \multicolumn{4}{c}{FACTIFY2}                                            \\
\hline
\textbf{Model}    & \textbf{Inputs} & \textbf{Support} & \textbf{Refute} & \textbf{NEI}   & \textbf{F1-macro} & \textbf{Support} & \textbf{Refute} & \textbf{NEI}   & \textbf{F1-macro} \\
\hline
Qwen-VL & mm\_claim       & 0.467            & 0.459           & \textbf{0.463} & 0.463             & 0.238            & 0.505           & 0.513          & 0.418             \\
Idefics2-8b       & mm\_claim       & \textbf{0.522}   & 0.535           & 0.399          & 0.485             & 0.427            & 0.516           & 0.471          & 0.471             \\
PaliGemma-3b      & mm\_claim       & 0.495            & 0.510           & 0.451          & 0.485             & 0.398            & 0.387           & 0.503          & 0.429             \\
\hline
Qwen-VL & mm\_claim+mm\_evidence    & 0.483            & 0.561           & 0.417          & 0.487             & \textbf{0.532}   & 0.443           & 0.469          & 0.481             \\
Idefics2-8b       & mm\_claim+mm\_evidence    & 0.501            & 0.572           & 0.429          & 0.501             & 0.339            & \textbf{0.674}  & 0.560          & \textbf{0.524}    \\
PaliGemma-3b      & mm\_claim+mm\_evidence    & 0.522            & \textbf{0.592}  & 0.444          & \textbf{0.519}    & 0.307            & 0.604           & \textbf{0.575} & 0.495            

\end{tabular}
}
\caption{Intrinsic fusion of VLM embeddings: Feed-forward neural classification with VLM embeddings}
  \label{tab:vlm-embeddings}
\end{table*}

\subsection{Intrinsic Fusion of VLM Embeddings}
\label{sec:exp_vlm}

In this experiment, we examined whether inherently multimodal models effectively utilize both text and image information. First, we extracted embeddings from selected VLMs and fed these vector representations into a feed-forward multi-class classifier. We extracted the last hidden states and applied mean pooling to each token's embedding. In other words, the extracted embedding size was \textit{(1, ntokens, ndim)}, where \textit{ntokens} is the number of tokens and \textit{ndim} is the dimension of each token embedding. Mean pooling provided a single embedding for each instance.

We provided two sets of inputs for extracting embeddings: \textit{mm\_claim} and \textit{mm\_evidence}. The \textit{mm\_claim} input consists of a claim and a corresponding image while the \textit{mm\_evidence} input consists of text evidence and an evidence image. For the second setting, we fed two input vectors to the classifier network: the \textit{mm\_claim} embedding and the \textit{mm\_evidence} embedding. This is because \textit{mm\_evidence} includes only the evidence representation - evidence image and evidence text - so we provided the claim information by feeding a second input to the classifier.

According to Table \ref{tab:vlm-embeddings}, the \textit{mm\_evidence} input setting improved F1-macro scores consistently for all models. This indicates that using both text and image evidence improved classification performance on both datasets. The results suggest that the selected VLMs effectively leverage information from evidence text and images on both the Mocheg and Factify2 datasets.

\subsection{Extrinsic Fusion of Language Model and Vision Encoder Embeddings}

\begin{table*}[th!]
\centering
\resizebox{\linewidth}{!}{
\begin{tabular}{ll|rrrr|rrrr}
                  &                                         & \multicolumn{4}{c}{MOCHEG}                                              & \multicolumn{4}{c}{FACTIFY2}                                            \\
\hline
\textbf{Model}    & \textbf{Inputs}                         & \textbf{Support} & \textbf{Refute} & \textbf{NEI}   & \textbf{F1-macro} & \textbf{Support} & \textbf{Refute} & \textbf{NEI}   & \textbf{F1-macro} \\
\hline
Qwen-7B+Vit-bigG  & claim+image                             & 0.472            & 0.533           & 0.438          & 0.481             & 0.520            & 0.854           & 0.514          & 0.629             \\
Mistral-7B+SigLIP & claim+image                             & 0.515            & 0.555           & \textbf{0.498} & 0.522             & 0.095            & 0.951           & \textbf{0.654} & 0.566             \\
Gemma-2b+SigLIP   & claim+image                             & 0.506            & 0.555           & 0.430          & 0.497             & 0.479            & 0.809           & 0.481          & 0.590             \\
\hline
Qwen-7B+Vit-bigG  & claim+claim\_image+text+text\_image & 0.486            & 0.577           & 0.413          & 0.492             & 0.398            & 0.788           & 0.558          & 0.581             \\
Mistral-7B+SigLIP & claim+claim\_image+text+text\_image & 0.503            & 0.574           & 0.407          & 0.495             & 0.580            & 0.607           & 0.362          & 0.516             \\
Gemma-2b+SigLIP   & claim+claim\_image+text+text\_image & 0.500            & 0.584           & 0.378          & 0.487             & 0.580            & 0.607           & 0.362          & 0.556             \\
\hline
Qwen-VL           & mm\_claim+mm\_image                     & 0.528            & 0.515           & 0.462          & 0.502             & 0.318            & 0.806           & 0.642          & 0.589             \\
Idefics2-8b       & mm\_claim+mm\_image                     & \textbf{0.555}   & 0.578           & 0.452          & \textbf{0.528}    & 0.437            & \textbf{0.982}  & 0.593          & \textbf{0.670}    \\
PaliGemma-3b      & mm\_claim+mm\_image                     & 0.551            & 0.453           & 0.390          & 0.465             & 0.606            & 0.583           & 0.000          & 0.396             \\
\hline
Qwen-VL           & mm\_text+mm\_image                      & 0.499            & \textbf{0.612}  & 0.431          & 0.514             & 0.519            & 0.812           & 0.530          & 0.620             \\
Idefics2-8b       & mm\_text+mm\_image                      & 0.526            & 0.541           & 0.458          & 0.509             & 0.319            & 0.825           & 0.547          & 0.564             \\
PaliGemma-3b      & mm\_text+mm\_image                      & 0.467            & 0.512           & 0.447          & 0.475             & \textbf{0.623}   & 0.681           & 0.001          & 0.435            
           
\end{tabular}
}
\caption{Extrinsic fusion of embeddings: Feed-forward neural classification with distinct text and image embeddings}
  \label{tab:fusion-embeddings}
\end{table*}

Separate embeddings were extracted for text and image information from the vision encoders and language models, respectively. Afterward, we performed mean pooling to obtain one-dimensional vector representations for each instance. For this experiment, we had four input setups:

\textbf{Input1 (claim+image):} The claim representation was taken from the language model and the corresponding image representation was taken from the vision transformer. 

\textbf{Input2 (claim+claim\_image+text+text\_image):} In addition to Input1, the evidence text representation was extracted from the language model and the evidence image representation was extracted from the vision transformer.

\textbf{Input3 (mm\_claim+mm\_image):} 
The embeddings extracted when the claim text is given to the VLM and the embeddings extracted when only the claim image is given were used separately.

\textbf{Input4 (mm\_text+mm\_image):} The embeddings extracted when all textual content is given to the VLM and the embeddings extracted when only the images are given were used separately.

Inputs, except Input2, had two separate text and image embeddings. Only the second setup had four embeddings: claim embedding, claim image embedding, text embedding, and text image embedding. After extracting the embeddings, we trained the proposed probing classifier as described in Section \ref{sec:probing_classifier} for multi-class veracity prediction. We extracted the embeddings for Input1 and Input2 using the selected multimodels' text and vision encoders that were also mentioned in Section \ref{sec:models}.

According to Table \ref{tab:fusion-embeddings}, Idefics2 with the third input setup outperformed the other models on both datasets. Note that Idefics2 also performed better in zero-shot evaluations which could indicate that the model might have encountered similar data during pre-training. Therefore, it may leverage its pre-training knowledge while processing these claims.

\subsection{Ablation Study}

\begin{table*}[th!]
\centering
\resizebox{\linewidth}{!}{
\begin{tabular}{lll|rrrr|rrrr}
\multicolumn{1}{c}{}               & \multicolumn{1}{c}{}  & \multicolumn{1}{c}{}               & \multicolumn{4}{c}{MOCHEG}                                                                                                                            & \multicolumn{4}{c}{FACTIFY2}                                                                                                                          \\
\hline
\textbf{Method}           & \textbf{Model}    & \textbf{Inputs} & \textbf{Support} & \textbf{Refute} & \textbf{NEI}   & \textbf{F1-macro} & \textbf{Support} & \textbf{Refute} & \textbf{NEI}   & \textbf{F1-macro} \\
\hline
\multirow{6}{*}{KNN} & Qwen-VL        & mm\_claim       & 0.253            & 0.433           & 0.235          & 0.307             & 0.422            & 0.025           & 0.485          & 0.311             \\
                     & Idefics2-8b    & mm\_claim       & 0.254            & 0.438           & 0.276          & 0.322             & 0.394            & 0.013           & 0.471          & 0.308             \\
                     & PaliGemma-3b   & mm\_claim       & 0.237            & 0.435           & 0.250          & 0.307             & 0.410            & 0.009           & 0.471          & 0.293             \\
                     & Qwen-VL        & mm\_claim+mm\_evidence    & 0.207            & 0.433           & 0.160          & 0.267             & 0.417            & 0.023           & 0.484          & 0.299             \\
                     & Idefics2-8b    & mm\_claim+mm\_evidence    & 0.206            & 0.450           & 0.122          & 0.259             & 0.405            & 0.016           & 0.477          & 0.296             \\
                     & PaliGemma-3b   & mm\_claim+mm\_evidence    & 0.150            & 0.457           & 0.148          & 0.252             & 0.401            & 0.017           & 0.471          & 0.296             \\
\hline
\multirow{6}{*}{SVM} & Qwen-VL        & mm\_claim       & 0.375            & 0.453           & 0.273          & 0.367             & 0.234            & 0.156           & 0.512          & 0.301             \\
                     & Idefics2-8b    & mm\_claim       & \textbf{0.432}   & 0.491           & \textbf{0.284} & \textbf{0.402}    & 0.268            & \textbf{0.238}  & 0.479          & 0.217             \\
                     & PaliGemma-3b   & mm\_claim       & 0.412            & 0.487           & 0.263          & 0.387             & 0.000            & 0.233           & \textbf{0.533} & \textbf{0.328}    \\
                     & Qwen-VL        & mm\_claim+mm\_evidence    & 0.380            & 0.490           & 0.233          & 0.368             & 0.583            & 0.046           & 0.023          & 0.320             \\
                     & Idefics2-8b    & mm\_claim+mm\_evidence    & 0.392            & 0.514           & 0.231          & 0.379             & \textbf{0.592}   & 0.187           & 0.181          & 0.255             \\
                     & PaliGemma-3b   & mm\_claim+mm\_evidence    & 0.383            & \textbf{0.521}  & 0.256          & 0.387             & 0.558            & 0.141           & 0.276          & 0.325            
\end{tabular}
}
\caption{Baseline classifiers' results}
  \label{tab:knn}
\end{table*}

Our feed-forward classifier, illustrated in Figure \ref{fig:ffclassifier}, consists of two sequential linear layers. The first layer resizes each input tensor to a "hidden size" before concatenating the tensors. We chose this approach because there was a significant difference between the image and text embedding sizes. By reshaping each tensor to the same size before concatenation, we aimed to utilize both types of information more effectively.

However, this approach has some limitations. If concatenation were performed before the first hidden layer, linear layers would be common for all models and input setups. In our approach, only the layers after concatenation are common so as the number of inputs increases, the number of learned parameters for the non-common layers also increases. Additionally, we did not validate the depth of the neural classifier and the network depth might be too shallow for the veracity detection task.

To assess whether the neural classifier effectively learns the intended task, we conducted an experiment using KNN and SVM classifiers with the same training embeddings as mentioned in Section \ref{sec:exp_vlm}. We set the number of neighbors (k), to seven which was decided after exploring consecutive values. Similarly, we trained SVM classifier with a linear kernel. As shown in Table \ref{tab:knn}, our approach outperformed the baselines on both datasets which implies that the proposed neural classifier leveraged the embeddings much better than the KNN and SVM classifiers on both datasets.
\section{Discussion}

First, we addressed RQ1 by conducting a zero-shot experiment to verify that multimodality improves performance depending on the dataset and model configuration, with models like Idefics-8b and LVLM4FV outperforming their text-only counterparts. Idefics2-8b benefits from image information while LVLM4V’s efficient prompting strategy further enhances verification accuracy.

Additionally, the proposed intrinsic fusion pipeline which utilizes VLM embeddings, outperformed the VLMs' base inference performance (see Table \ref{tab:vlm-inference} and Table \ref{tab:vlm-embeddings}). The only exception was the Idefics2 model on the Mocheg dataset, which had a 0.517 F1-macro inference score while the classifier achieved only a 0.501 F1-macro score. Since the probing classifier has only two layers, it might be too shallow for this dataset and model. Note that the primary goal of this study is not to achieve state-of-the-art scores for the selected datasets. Instead, we aim to evaluate whether recent VLMs improve performance on the fact-checking problem through multimodality or if fusing externally the information from distinct models achieves superior results.

Secondly, we addressed RQ2 by assessing how VLMs leverage text and image information. According to the results, for Idefics2-8b and Qwen-VL, multimodal embeddings were outperformed by discrete models (see Table \ref{tab:vlm-embeddings} and Table \ref{tab:fusion-embeddings}). In other words, extracting separate embeddings resulted in higher F1-macro scores across all models. To be more specific, on the Mocheg dataset, the highest F1-macro scores for Qwen-VL and Idefics-8b were 0.514 and 0.528 respectively. Similarly, on the Factify2 dataset, the highest F1-macro scores were 0.629, 0.670 and 0.590 respectively. Although the best results were achieved with different input setups, for all of the best results, we extracted separate text and image embeddings. In contrast, when embeddings were extracted from inherently multimodal VLMs (as shown in Table \ref{tab:vlm-embeddings}), the maximum F1-macro scores were lower except PaliGemma-3b on Mocheg dataset. This indicates that for the given evaluation framework, using discrete text and image embeddings yielded higher F1-macro scores.

Besides, RQ3 was addressed by conducting an ablation study to examine how the proposed classifier leverages embeddings against KNN and SVM baselines. According to our evaluations, the proposed classifier utilized the extracted embeddings significantly better than the baseline approaches.

Finally, on the Mocheg dataset, the selected models struggle more on "not enough info" cases, as their lowest success rates, even in the best settings, were consistently associated with this class. This may be due to class relabeling, where the authors of the Mocheg dataset reannotated the "Mixture," "Unproven," and "Multiple" cases as "Not Enough Info" which may lead to confusion for the models. In contrast, on the Factify2 dataset, the trained classifier was more successful in distinguishing fake claims compared to other classes. This could be linked to the difference of data domains, as the genuine news was sourced from news agencies while fake claims were crawled from fact-checking sites and satirical articles.
\section{Conclusion}

In this study, we utilize VLMs for multimodal fact-checking and propose a probing classifier-based approach. The proposed pipeline extracts embeddings from the last hidden layer of selected VLMs and fuses multimodal embeddings (extrinsic or intrinsic) into a simple feed-forward neural network for multi-class veracity classification. The experiments show that employing a probing classifier is more effective than the base VLM performance and extrinsic fusion usually outperforms the intrinsic fusion for the proposed approach. As future work, we plan to employ VLMs as assistants rather than as primary fact-checkers. To be more specific, the VLM can be used as an assistant that reviews the given text and image and returns a summary or justification to guide the text-only model for the fact-checking task. Since the LLMs are prone to hallucination and their accuracy depends on the quality of their training data which may be outdated or biased, incorporating knowledge grounding could be a more reliable strategy for real-world deployment.

\section{Limitations}

We tested a limited number of models which may not fully capture the variability across different models and configurations. Additionally, the evaluations were performed on English datasets, restricting the assessment of multilingual capabilities. Furthermore, there is a potential risk that some dataset instances may overlap with the training data of the VLMs which could bias the evaluation results.

Moreover, while extracting embeddings from the selected VLMs and corresponding LLMs, we encountered some computational overhead. More specifically, for some claims, the evidence field exceeded the sequence length of the models or could not fit within our memory constraints. Therefore, we cropped the evidence fields for such instances. Furthermore, while LLMs and VLMs are prone to hallucination, we did not perform any analysis on this phenomenon within the scope of this study. 

\section*{Acknowledgments}

This research is supported by the Scientific and Technological Research Council of Turkey (TUBITAK, Prog: 2214-A) and the German Academic Exchange Service (DAAD, Prog: 57645447). We would like to thank the anonymous reviewers for their suggestions to improve the study. We also appreciate METU-ROMER and the University of Tübingen for providing the computational resources.

This project is partially supported by METU with grant no ADEP-312-2024-11484. Parts of this research received the support of the EXA4MIND project, funded 
by the European Union´s Horizon Europe Research and Innovation Programme, 
under Grant Agreement N° 101092944. Views and opinions expressed are 
however those of the author(s) only and do not necessarily reflect those 
of the European Union or the European Commission. Neither the European 
Union nor the granting authority can be held responsible for them.

\bibliography{custom}

\appendix

\appendix
\begin{table*}[th!]
\centering
\resizebox{\linewidth}{!}{
\begin{tabular}{ll|rrrr|rrrr}
                   &                          & \multicolumn{4}{c}{\textbf{MOCHEG}}                                                    & \multicolumn{4}{c}{\textbf{FACTIFY2}}                                                  \\
\textbf{Embedding} & \textbf{Input}           & \textbf{Batch size} & \textbf{Learning rate} & \textbf{Hidden size} & \textbf{Dropout} & \textbf{Batch size} & \textbf{Learning rate} & \textbf{Hidden size} & \textbf{Dropout} \\
\hline
Qwen-VL            & mm\_claim                & 32                  & 0.01                   & 128                  & 0.1              & 64                  & 0.001                  & 128                  & 0.1              \\
idefics2-8b        & mm\_claim                & 32                  & 0.01                   & 256                  & 0.05             & 32                  & 0.0001                 & 128                  & 0.1              \\
PaliGemma-3b       & mm\_claim                & 32                  & 0.01                   & 512                  & 0.05             & 64                  & 0.0001                 & 128                  & 0.05             \\
\hline
Qwen-VL            & mm\_claim + mm\_evidence & 64                  & 0.01                   & 256                  & 0.05             & 32                  & 1E-05                  & 256                  & 0.05             \\
idefics2-8b        & mm\_claim + mm\_evidence & 64                  & 0.01                   & 512                  & 0.1              & 32                  & 0.001                  & 256                  & 0.1              \\
PaliGemma-3b       & mm\_claim + mm\_evidence & 64                  & 0.001                  & 256                  & 0.1              & 64                  & 1E-05                  & 512                  & 0.1              \\
\hline
Qwen-7B+Vit-bigG   & input1                   & 128                 & 0.01                   & 512                  & 0.1              & 32                  & 0.001                  & 128                  & 0.1              \\
Mistral-7B+SigLIP  & input1                   & 64                  & 0.001                  & 512                  & 0.1              & 128                 & 0.001                  & 256                  & 0.2              \\
Gemma-2b+SigLIP    & input1                   & 64                  & 0.01                   & 512                  & 0.1              & 128                 & 0.001                  & 128                  & 0.1              \\
\hline
Qwen-7B+Vit-bigG   & input2                   & 32                  & 0.001                  & 256                  & 0.4              & 64                  & 0.001                  & 128                  & 0.1              \\
Mistral-7B+SigLIP  & input2                   & 64                  & 0.01                   & 512                  & 0.1              & 64                  & 0.001                  & 256                  & 0.4              \\
Gemma-2b+SigLIP    & input2                   & 64                  & 0.001                  & 512                  & 0.2              & 64                  & 0.001                  & 256                  & 0.4              \\
\hline
Qwen-VL            & input3                   & 32                  & 0.001                  & 512                  & 0.2              & 128                 & 0.001                  & 512                  & 0.1              \\
Idefics2-8b        & input3                   & 128                 & 0.001                  & 512                  & 0.1              & 128                 & 0.01                   & 512                  & 0.1              \\
PaliGemma-3b       & input3                   & 64                  & 0.001                  & 256                  & 0.1              & 64                  & 0.001                  & 256                  & 0.2              \\
\hline
Qwen-VL            & input4                   & 64                  & 0.001                  & 512                  & 0.1              & 128                 & 0.001                  & 128                  & 0.4              \\
Idefics2-8b        & input4                   & 128                 & 0.001                  & 128                  & 0.4              & 128                 & 0.001                  & 128                  & 0.1              \\
PaliGemma-3b       & input4                   & 64                  & 0.001                  & 256                  & 0.2              & 32                  & 0.001  & 512                  & 0.4           
\end{tabular}
}
\caption{Parameter settings for the best models}
\label{tab:val_params}
\end{table*}

\begin{table}[ht!]
\centering
\resizebox{\linewidth}{!}{%
\begin{tabular}{l|rr}
\multicolumn{1}{c}{\textbf{Model}} & \multicolumn{1}{c}{\textbf{Mocheg (1655)}} & \multicolumn{1}{c}{\textbf{Factify2 (7273)}} \\
\hline
Qwen-7B                            & 1366 (82.5\%)                              & 4335 (59.6\%)                                \\
Mistral-7B                         & 1361 (82.2\%)                              & 5756 (79.1\%)                                \\
Gemma-2B                           & 1617 (97.7\%)                              & 6136 (84.4\%)                                \\
\hline
Qwen-VL                            & 1646 (99.5\%)                              & 6483 (89.1\%)                                \\
Idefics2-8b                        & 1653 (99.9\%)                              & 5873 (80.7\%)                                \\
PaliGemma-3b                       & 320 (19.3\%)                               & 91 (1.2\%)                             
\end{tabular}
}
\caption{Zero-shot response frequencies}
\label{tab:zs_frequency}
\end{table}

\section{Hyperparameter Values for the Best Models}
\label{sec:appendix_hyperparam}

We set the number of epochs to 20, enabling early stopping with the patience of 5 and monitoring the validation loss. We used the Adam optimizer in combination with a cosine scheduler, employing a warm-up ratio of 0.05. Moreover, we adjusted the cross-entropy loss weight of the neural network according to the inverse class ratios. In this way, the classifier was penalized more for the misclassifications of the minority classes. 

We performed a grid search to explore the following parameter space for the results given in Table \ref{tab:vlm-embeddings} and Table \ref{tab:fusion-embeddings}:

\noindent{\textit{learning rate}}: \{ 0.00001, 0.0001, 0.001, 0.01, 0.1\}, 
\noindent{\textit{batch size}}: \{32, 64, 128\}, 
\noindent{\textit{hidden size}} (h in Figure \ref{fig:ffclassifier}): \{128, 256, 512 \} and  
\noindent{\textit{dropout}}: \{0.05, 0.1, 0.2, 0.4\}.

The parameter settings for the best results are detailed in Table \ref{tab:val_params}.

\section{Zero-shot Model Response Frequencies}
\label{sec:appendix_zs}
We used the prompt template shown in Figure \ref{fig:prompt} for all models in the zero-shot inference experiments. We expected the models’ responses to contain either "supported," "refuted," or "not enough info." If a model’s response did not contain these labels, we ignored those instances. Additionally, we observed that PaliGemma consistently responded with "sorry, as a base VLM I am not trained to answer this question," which could be due to injected policies. The frequencies of considered cases for each model (with percentages in parenthesis) are given in Table \ref{tab:zs_frequency}.

\section{Fine-tuning Parameter Settings}
\label{sec:appendix_ft}
We employed QLoRA \cite{dettmers2024qlora} adapter on top of attention weight matrices and fine-tuned only the LoRA \cite{hu2022lora} adapters for 3 epochs. The batch size was set to 2 with an initial learning rate of 2e-5 using a cosine scheduler and the Adam optimizer. We used the checkpoint with the lowest validation loss. Additionally, we set warm up to 0.02, gradient accumulation to 4 and evaluated on validation set 10 times during fine-tuning. We set the rank of matrices for LoRA adapters to 16, the scaling factor (lora\_alpha) to 16 and the dropout rate for the adapters to 0.05. Besides, 16-bit mixed precision, bfloat16, was employed for memory efficiency and faster fine-tuning.


\end{document}